\documentclass[lettersize,journal]{IEEEtran}
\usepackage{amsmath,amsfonts}
\usepackage{algorithmic}
\usepackage{algorithm}
\usepackage{array}
\usepackage[caption=false,font=normalsize,labelfont=sf,textfont=sf]{subfig}
\usepackage{textcomp}
\usepackage{stfloats}
\usepackage{url}
\usepackage{verbatim}
\usepackage{graphicx}
\usepackage{cite}
\usepackage{tabularx}
\usepackage{float}
\usepackage{caption}
\usepackage{bm}
\usepackage{multirow}
\usepackage{amssymb}
\usepackage{multirow}
\usepackage{slashed}

\usepackage[marginal]{footmisc}

\usepackage{xcolor}

\begin{document}
	
	\title{KKA: Improving Vision Anomaly Detection through Anomaly-related Knowledge from Large Language Models}
	

	\author{Dong~Chen, Zhengqing~Hu, Peiguang~Fan, Yueting~Zhuang,~\IEEEmembership{Senior~Member,~IEEE}, Yafei~Li, Qidong~Liu, Xiaoheng Jiang$^{*}$, Mingliang Xu$^{*}$\thanks{*Both Xiaoheng Jiang and Mingliang Xu are corresponding authors.},~\IEEEmembership{Member,~IEEE}   
	\IEEEcompsocitemizethanks{\IEEEcompsocthanksitem
		D. Chen, Z. Hu, P. Fan, Y. Li, Q. Liu, X. Jiang, M. Xu are with the School of Computer and Artificial Intelligence of  Zhengzhou University, Zhengzhou, China. Engineering Research Center of Intelligent Swarm Systems, Ministry of Education, Zhengzhou, China. National Supercomputing Center In Zhengzhou , Zhengzhou, China
(E-mail: chendongcs@zju.edu.cn, superminih1998@gs.zzu.edu.cn, fanpeiquang@gs.zzu.edu.cn, ieyfli@zzu.edu.cn, ieqdliu@zzu.edu.cn, jiangxiaoheng@zzu.edu.cn, iexumingliang@zzu.edu.cn). Y. Zhuang is with Zhejiang University, Hangzhou, China (E-mail: yzhuang@zju.edu.cn).
}
}

	
	\maketitle
	
	\begin{abstract}
Vision anomaly detection, particularly in unsupervised settings, often struggles to distinguish between normal samples and anomalies due to the wide variability in anomalies. Recently, an increasing number of studies have focused on generating anomalies to help detectors learn more effective boundaries between normal samples and anomalies. However, as the generated anomalies are often derived from random factors, they frequently lack realism. Additionally, randomly generated anomalies typically offer limited support in constructing effective boundaries, as most differ substantially from normal samples and lie far from the boundary. To address these challenges, we propose Key Knowledge Augmentation (KKA), a method that extracts anomaly-related knowledge from large language models (LLMs). More specifically, KKA leverages the extensive prior knowledge of LLMs to generate meaningful anomalies based on normal samples. Then, KKA classifies the generated anomalies as easy anomalies and hard anomalies according to their similarity to normal samples. Easy anomalies exhibit significant differences from normal samples, whereas hard anomalies closely resemble normal samples. KKA iteratively updates the generated anomalies, and gradually increasing the proportion of hard anomalies to enable the detector to learn a more effective boundary. Experimental results show that the proposed method significantly improves the performance of various vision anomaly detectors while maintaining low generation costs. The code for CMG can be found at https://github.com/Anfeather/KKA.
	\end{abstract}
	
	
	\section{Introduction}
	\label{Introduction}

Unsupervised anomaly detection is one of the most important fields of anomaly detection, where no prior information on anomalies is available, while there are normal samples for reference \cite{chen2024improvingg,zhou2022pull,chen2020simple,tack2020csi,chen2023cross}.
Unsupervised methods often struggle to achieve satisfactory results due to the difficulty of learning effective boundaries between anomalies and normal samples. This challenge is particularly pronounced for visual data, where even images within the same category can exhibit significant variations, further complicating anomaly detection.
To address this issue, some generation-based anomaly detection methods have been proposed. However, anomalies generated by these methods often result from random factors and deviate significantly from real-world scenarios \cite{liu2023simplenet}. Furthermore, excessive randomness in the generated anomalies provides limited guidance for defining boundaries, resulting in wasted computational resources and minimal improvement.

\begin{figure}[t]
	\centering
	\includegraphics[scale=0.45]{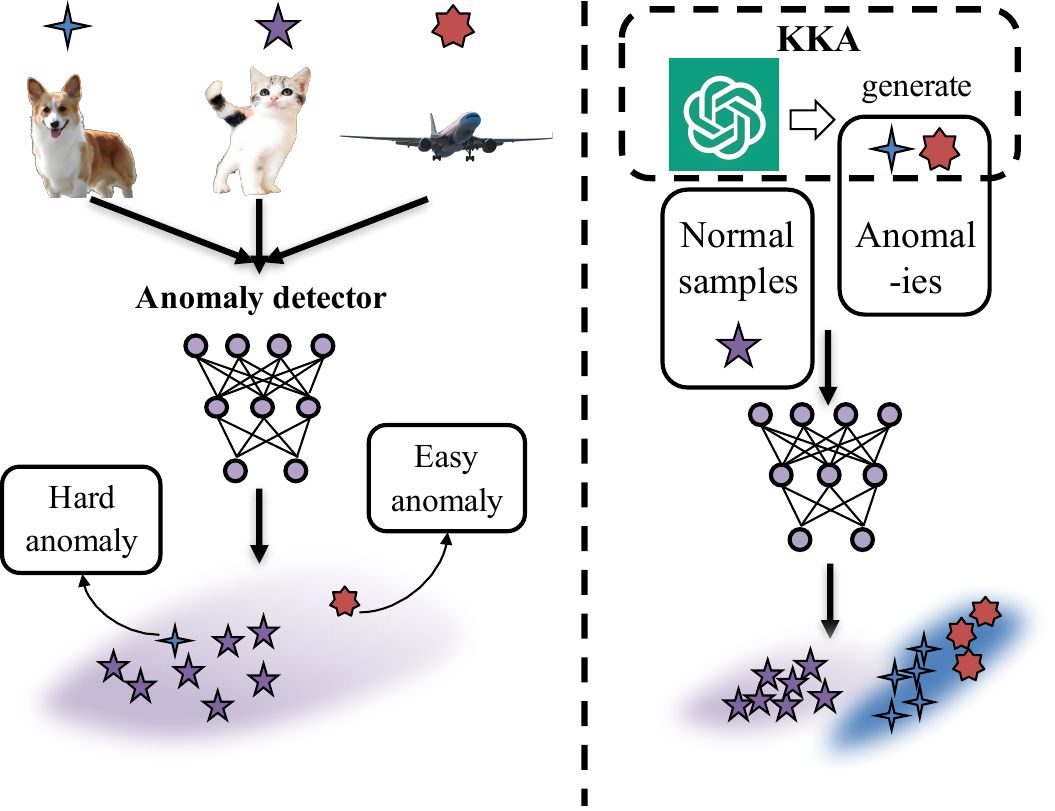}
	\caption{Unsupervised anomaly detection and anomaly detection utilizing anomaly knowledge derived from LLMs.}
	\label{Fig: U_VS_S}
\end{figure}
As depicted in the left part of Figure \ref{Fig: U_VS_S}, in the context of unsupervised vision anomaly detection, anomalies such as airplanes significantly deviate from normal samples like cats. Consequently, the detector can more easily differentiate anomalies from normal samples, which we refer to as ``easy anomalies". 
Conversely, when anomalies, such as dogs, resemble normal samples, the detector struggles to learn an effective boundary between normal samples and anomalies. We refer to such anomalies as hard anomalies.
Hard anomalies, due to their close resemblance to normal samples, are often more difficult to obtain but have a stronger influence on the detector's learned boundary.

Recently, large language models (LLMs) like GPT-3.5 \cite{ouyang2022training} and GLM-4 \cite{glm2024chatglm} have garnered widespread attention due to their vast pretrained knowledge and powerful abilities in understanding and generation \cite{zhao2023survey}. 
Based on this, some researchers have attempted to employ LLMs to distinguish anomalies \cite{elhafsi2023semantic, zanella2024harnessing}.
While LLMs offer a promising research direction for anomaly detection, their substantial parameter scale presents challenges for deployment on most edge devices. This limitation is especially problematic in vision anomaly detection scenarios, such as video surveillance systems that rely on edge computing \cite{patrikar2022anomaly,yu2022edge,ngo2020adaptive}.
Besides, previous studies \cite{chen2024data} indicate that the collaboration between large and small models can match, or even exceed, the performance of large models alone.
Thus, we focus on improving the performance of the small vision detector by extracting anomaly-related knowledge from LLMs, and we refer to the knowledge about hard anomalies in LLMs as key knowledge. 

We propose Key Knowledge Augmentation (KKA), which leverages the anomaly-related knowledge from LLMs, particularly key knowledge, to generate plausible anomalies tailored to various anomaly detection scenarios.
Specifically, KKA first randomly samples normal samples and designs prompts to enable LLMs to generate anomaly-related knowledge, and the knowledge is then used to create various anomaly images for detector training. To further enhance the detector’s performance, KKA trains a confusion evaluator to identify hard anomalies among the generated images. Besides, the text descriptions of hard anomalies will be used to iteratively train LLMs to extract more key knowledge.
As shown in the right part of Figure \ref{Fig: U_VS_S}, with augmented key knowledge from LLMs, detector can learn a clearer boundary for normal samples and anomalies.


The main contributions of this paper can be summarized as follows:
\begin{itemize}
	\item We analyzed the issues of existing generation-based methods from the perspective of generated anomalies.
	\item We propose extracting anomaly-related knowledge, particularly key knowledge, from LLMs and iteratively optimizing the detector's performance with a specially designed LLM.
	\item We extensively evaluate the proposed Key Knowledge Augmentation (KKA) on various datasets and achieve excellent results. For example, on CIFAR-100, KKA improves the AUC of the recent generation-based method SimpleNet from $74.62\%$ to $84.04\%$, while generating only about $5\%$ of the samples produced by SimpleNet.
\end{itemize}

\section{Related work}
\label{sec:RW}

Anomaly detection has a very wide range of applications, such as detecting defects in industrial products \cite{liu2024deep,roth2022towards,siegel2020industrial}, financial fraud \cite{hilal2022financial,huang2018codetect}, as well as video surveillance \cite{zhang2020normality,li2020spatial,chang2021contrastive}. No matter in which scenario, anomalies are sparse and unpredictable \cite{pang2021deep}.

\subsection{Unsupervised Anomaly Detection with Generated Anomalies}
Unsupervised anomaly detection is one of the most important fields of anomaly detection. When training a detector using only normal samples, distinguishing anomalies with diverse distributions becomes challenging \cite{zong2018deep,sehwag2021ssd}. Particularly in the case of images, samples from the same class may exhibit significant variation, which can obscure the boundary between normal samples and anomalies in the latent space \cite{chen2024improvingg}. 
Consequently, a growing number of studies focus on improving detector's performance by generating additional samples.
CMDA \cite{chen2023cross} increases the sample size by interpolating between text and images, thereby enhancing the detector's discriminative capacity.
SimpleNet \cite{liu2023simplenet} generates anomalies by introducing Gaussian noise into the image feature space and trains the detector at the feature level.
ReContrast \cite{guo2024recontrast} combines feature reconstruction and contrastive learning by using two different encoders to generate distinct versions of the same sample, thereby enhancing the detector's ability to discriminate across diverse distributions.
\subsection{Anomaly Detection with LLMs}
Motivated by the advanced comprehension, powerful generation capabilities, and vast prior knowledge of LLMs, recent studies have increasingly explored integrating anomaly detection with LLMs.
Semantic Anomaly Detection \cite{elhafsi2023semantic} introduces a monitoring framework that utilizes LLMs to detect semantic anomalies in vision-based policies, aiming to identify failures at the system level.
LAnguage-based VAD (LAVAD) \cite{zanella2024harnessing} employs LLMs to detect anomalies through analysis of scene descriptions. 
AnomalyLLM \cite{liu2024large} effectively identifies anomalies by detecting significant discrepancies between the features of LLMs and those of a smaller anomaly detector. Although the aforementioned methods have improved anomaly detectors with the help of LLMs, the large parameter scale of LLMs limits the applicability of these methods \cite{patrikar2022anomaly,yu2022edge,ngo2020adaptive}.
On the other hand, prior studies indicate that smaller specialized models can achieve performance comparable to, or even exceeding, that of general large models on specific distributions \cite{chen2024data,chen2024improving}. Consequently, this paper focuses on employing LLMs to enhance small anomaly detectors.

	\section{Method}
	\label{sec: method}

In this section, we present the proposed Key Knowledge Augmentation (KKA). First, we introduce the symbols and their corresponding meanings, as outlined in Table \ref{Symbol_Definitions}.

\begin{table}[h]
	\caption{Symbol Definitions.}
	\label{Symbol_Definitions}
	\begin{tabularx}{\linewidth}{|c| X<{\centering}|  }
		\hline Symbol&  Definitions\\
		\hline
		$\bullet^i$& image.\\
		$\bullet^t$& text description.\\
		$X_n$ &dataset of normal samples.\\
		$X_a$ &dataset of anomalies.\\
		$x_n$ & $x_n\in$ $X_n$ , normal sample.\\
		$x_{a}$ & generated anomaly.\\
		$x_{ea}$ & generated easy anomaly.\\
		$x_{ha}$ & generated hard anomaly.\\
		$\phi$	&	feature extractor.\\
		$\mathcal{Z}$ & embedding space learned by detector.\\
		$\theta_C$	&confusion evaluator.\\
		$\theta_L$ & LLM.\\
		$\theta_S$ & text-to-image model.\\
		\hline
	\end{tabularx}
\end{table}

\subsection{Preliminary}
Due to the sparsity and diversity of anomalies, anomaly detection methods, specifically unsupervised anomaly detection, often train a feature extractor $\phi$ with normal samples $X_n$. During testing, the detector calculates the anomaly score in the latent space learned by $\phi$ with a distance metric, such as Mahalanobis distance, to discriminate normal samples and anomalies \cite{chen2023cross, chen2024improvingg}:
\begin{equation}
s_k = (z-\mu_k)^T\Sigma_{k}^{-1}(z-\mu_k), \quad z=\phi(x)
\label{EQ: anomaly score}
\end{equation}
where $x\in X_n$ is the normal sample, $z\in \mathcal{Z}$ is the output of $\phi$, $k$ denotes different clusters in the latent space, $\mu_k$ and $\Sigma_k$ are the sample mean and sample covariance, respectively.

When anomalies deviate significantly from the normal distribution (i.e., easy anomalies, $x_{ea}$), the detector can effectively distinguish between normal samples and anomalies with Eq.\ref{EQ: anomaly score}. However, when normal samples and anomalies are more similar (i.e., hard anomalies, $x_{ha}$), the performance of unsupervised anomaly detection is considerably reduced.

\subsection{Key Knowledge Augmentation}
\begin{figure*}[t]
	\centering
	\includegraphics[scale=0.62]{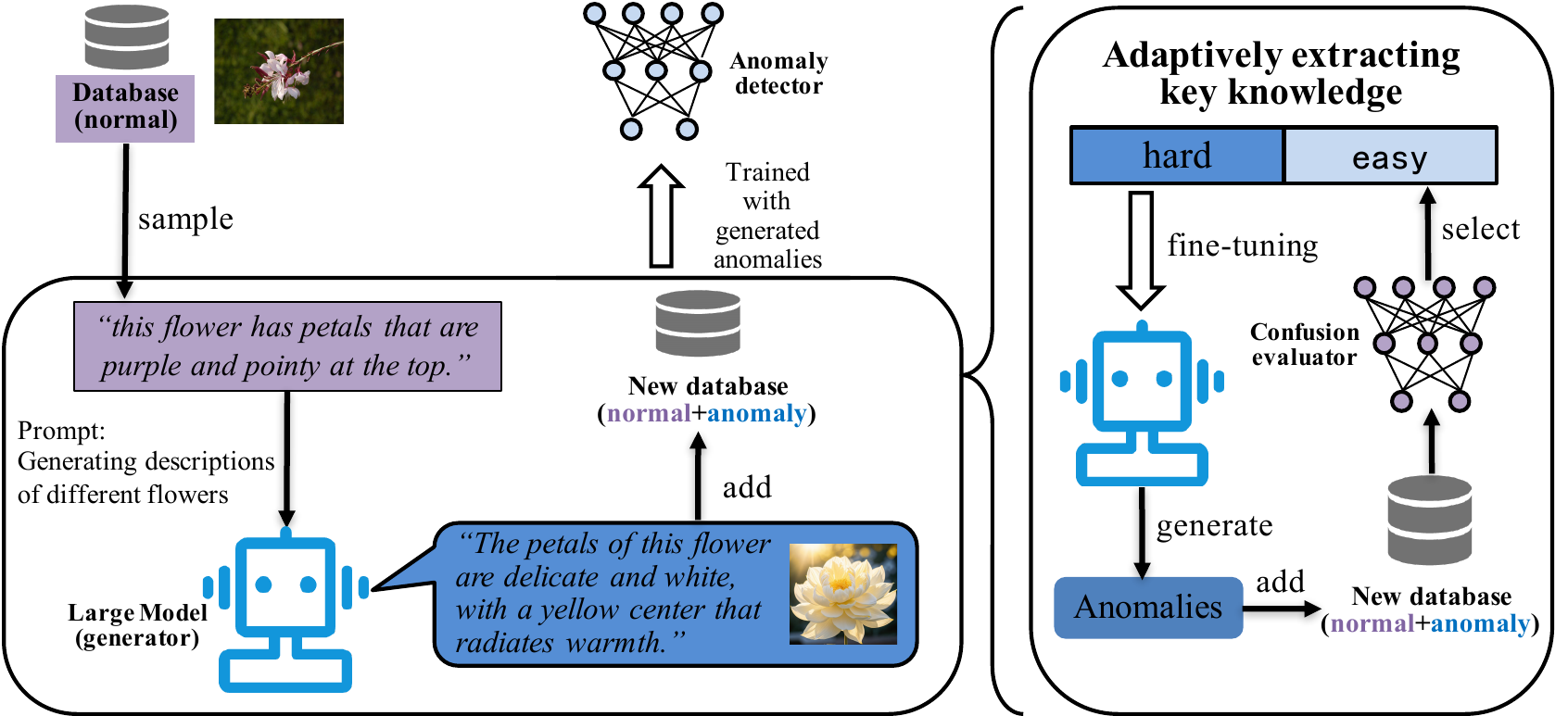}
	\caption{The overview of KKA. The purple dashed box indicates training the detector using only normal samples, while the blue dashed box indicates training the detector with both normal and generated anomalies.}
	\label{Fig: KS}
\end{figure*}

We propose leveraging the extensive prior knowledge of LLMs to predict anomalies, especially those closely resembling normal samples, thereby enhancing the performance of anomaly detectors. Specifically, we introduce Key Knowledge Augmentation (KKA), and the proposed method is illustrated in Figure \ref{Fig: KS}.

We use text to represent the knowledge within LLMs, and the knowledge of normal samples in the training data can be expressed as $X_n^t$. 
KKA first selects text descriptions that can represent the normal distributions from $X_n^t$:
\begin{equation}
\overline{X_n^t}=[x_{n1}^{t},x_{n2}^{t},...,x_{nK}^{t}] = \underset{x}{\arg \min }[s_1,s_2,...,s_K]
\label{Eq: selected_sample}
\end{equation}
where $K$ denotes the number of clusters of normal images, $s$ is the anomaly score calculated by Eq.\ref{EQ: anomaly score}, $[x_{n1}^{t},x_{n2}^{t},...,x_{nK}^{t}]$ are text descriptions of selected image list that best fits the normal distributions.

Then, KKA generates textual anomaly $x_a^t$ with different prompts according to the normal samples. For example, in Figure \ref{Fig: KS}, the normal samples are ``Gaura lindheimeri Engelm. \& A. Gray" (a type of flower), and the part of the prompt for generating anomalies is:

\textit{``Please generate new sentences that describe flowers. The sentences should follow the exact structure and style of the example below: $\overline{X_n^t}$".}

The textual anomaly $x_{a}^t$ generated by LLMs can be formatted as:
\begin{equation}
p_{\theta_L}(x_{a}^t|\overline{X_n^t})=\prod_{i}^{}p_{\theta_L}(y_i|\overline{X_n^t},y_{1:i-1})
\label{Eq: pseudo anomalies}
\end{equation}
where LLMs $\theta_L$ generates a current token $y_i$ based on a context of the previous $i -1$ tokens, and prompts. 

Thereafter, we can get anomaly image $x_{a}^i$ with textual anomaly $x_{a}^t$ and a text-to-image model such as Stable Diffusion $p_{\theta_S}$ \cite{rombach2022high}:
\begin{equation}
x_{a}^i \leftarrow p_{\theta_S}(x_{a}^i|x_{a}^t)
\label{Eq: t2i}
\end{equation}
where $x_{a}^i$ is the generated anomaly image corresponding to $x_{a}^t$.

By running Eq.\ref{Eq: t2i} multiple times, we can obtain an anomaly image dataset $X_a^i$.
Intuitively, the detector trained with $X_n^i$ and $X_a^i$ is already more effective in distinguishing between normal images and anomaly images, compared to unsupervised methods. However, anomalies generated by Eq.\ref{Eq: pseudo anomalies} and Eq.\ref{Eq: t2i} include numerous easy anomalies, $x_{ea}^i$, since hard anomalies, $x_{ha}^i$, must meet specific distribution conditions and demonstrate similarity to normal images.

In order to get more hard anomalies $x_{ha}^i$, KKA follows Deep SAD \cite{ruffdeep} to train a confusion evaluator with $X_n^i$ and $X_{a}^i$, which learns a transformation that minimizes the volume of a data-enclosing hypersphere in the latent space $\mathcal{Z}$:
\begin{equation}
\min _{\mathcal{W}} \frac{1}{M} \sum_{j=1}^M\left(\left\|E(\theta_C\left({x})\right)-\boldsymbol{c}\right\|^2\right)^{\tilde{y}_j}+\frac{\lambda}{2} \left\|\mathcal{W}\right\|_F^2,\lambda>0
\label{Eq: loss}
\end{equation}
where $\mathcal{W}$ is the set of weights of $\theta_C$, $M$ is the number of anomalies, $c$ is the predetermined center of hypersphere. Eq.\ref{Eq: loss} tends to project normal samples near the hypersphere center $c$, while pushing away anomalies from $c$.

Then, KKA calculates the distance from each sample to the center $c$ in the latent space, and selects anomalies whose distance to $c$ is shorter than that of certain normal samples. Such process can be formatted as:
\begin{equation}
x_{ha}^i \leftarrow DF(\theta_C(x_{a}^i ),c) < max(DF(\theta_C(x_{n}^i ),c))
\label{Eq: which is hard}
\end{equation}
where $DF$ is the distance function. Overall, Eq.\ref{Eq: which is hard} aims to identify hard anomalies that are easily confused with normal samples, while samples that do not satisfy Eq.\ref{Eq: which is hard} are considered easy samples.

To extract more key knowledge about hard anomalies from LLMs, KKA follows the Direct Preference Optimization (DPO) \cite{rafailov2024direct} to fine-tune LLMs:
\begin{equation}
\begin{aligned}
&\mathcal{L}\left(\theta_L^{'}, \theta_L\right)=\\&-{E}_{\left(x_n^{t}, x_{ea}^t, x_{ha}^t\right) }\\&\left[\log \sigma\left(\beta \log \frac{p_{\theta_L^{'}}\left(x_{ha}^t \mid x_n^t\right)}{p_{\theta_L}\left(x_{ha}^t \mid x_n^t\right)} - \beta \log \frac{p_{\theta_L^{'}}\left(x_{ea}^t \mid x_n^t\right)}{p_{\theta_L}\left(x_{ea}^t \mid x_n^t\right)}\right)\right]
\end{aligned}
\label{Eq: DPO}
\end{equation}
where $\sigma$ is the sigmoid function, $\theta_L^{'}$ is the fine-tuning LLM, $\theta_L$ is the original LLM.

After fine-tuning, we can extract more key knowledge from LLMs, and generate more hard anomalies by:
\begin{equation}
\begin{aligned}
&p_{\theta_L^{'}}(\overline{x_{a}^t}|x_n^t)=\prod_{i}^{}p_{\theta_L^{'}}(y_i|x_n^t,y_{1:i-1})\\
&\overline{x_{a}^i} \leftarrow p_{\theta_S}(x_{a}^i|x_{a}^t), N_{x_{ha}^i}({\overline{x_{a}^i}})>N_{x_{ha}^i}({x_{a}^i})
\label{Eq: hard_pseudo anomalies}
\end{aligned}
\end{equation}
where $N_{x_{ea}^i}(\cdot)$ denotes the number of hard anomaly images. 

In each iteration of updating anomaly dataset, KKA can get a new anomaly dataset $X_a^i$ with $\overline{x_{a}^i}$. To accommodate the varying needs of detectors for easy and hard anomalies in different scenarios, we divide KKA into KKA (add) and KKA (rep). KKA (add) continuously increases the number of anomalies (gradually increase the proportion of hard anomalies), while KKA (rep) maintains a constant number of anomalies by continuously replacing easy anomalies with harder ones. The whole process of KKA is summarized in Algorithm \ref{alg1}.

\begin{algorithm}
	\caption{Training process of KKA}
	\label{alg1}
	\begin{algorithmic}[1]
		\STATE	\textbf{Input}: Normal images $X_n^i$ and corresponding descriptions $X_n^t$.
		\STATE Select representative normal samples $\overline{X_n^t}$ through Eq.\ref{Eq: selected_sample}.
		\STATE Generate anomaly descriptions $x_{a}^t$ and anomaly images $x_{a}^i$ by Eq.\ref{Eq: pseudo anomalies} and Eq.\ref{Eq: t2i}, respectively.
		\STATE Construct anomaly dataset $X_a^i$ with $x_{a}^i$.
		\FOR{}
		\STATE  Train confusion evaluator $\theta_C$ by Eq.\ref{Eq: loss}.
		\STATE  Select hard anomalies $x_{ha}^i$ by Eq.\ref{Eq: which is hard}.
		\STATE  Fine-tune LLMs by Eq.\ref{Eq: DPO}.
		\STATE Update anomaly dataset $X_a^i$ by Eq.\ref{Eq: hard_pseudo anomalies}.
		\ENDFOR
		\STATE Return $X_a^i$
	\end{algorithmic}
	
\end{algorithm}

	\begin{table*}[ht]
		\centering
	\begin{tabularx}{\linewidth}{|c| X<{\centering}|  X<{\centering}| X<{\centering}|}
			\hline
			Dataset & Methods & AUC   &Number of generated anomalies\\
			\hline
			\multirow{11}{*}{CIFAR-100} 
			& SSD & $60.85\pm2.72\%$ &$0$\\
			&CMDA & $62.90\pm1.88\%$ & $375,000$ \\
			&SimpleNet  & $74.62\pm1.12\%$&$20,000$ \\
			&ReContrast (No pre-training)& $66.06 \pm 1.35\%$ & $1,600$ \\
			&ReContrast& $86.56 \pm 0.84\%$ & $1,600$ \\
			\cline{2-4}
			&SAD+KKA (add) & $72.71\pm0.98\%$ & $1,200$ \\
			&SAD+KKA (rep) & $72.26\pm1.19\%$ & $1,000$\\
			\cline{2-4}
			&SimpleNet+KKA (add)   & $81.97 \pm 1.67\%$&$20,000 (1,200)$ \\
			&SimpleNetSAD+KKA (rep)  & $84.04 \pm 2.64\%$&$20,000 (1,000)$ \\
			\cline{2-4}
			&ReContrast+KKA (add)   & $87.99 \pm 0.69\%$&$1,600+1,200$ \\
			&ReContrast+KKA (rep)  & $87.96 \pm 0.84\%$&$1,600+1,000$ \\
			\hline
			\multirow{11}{*}{Oxford-102} 
			& SSD & $72.74 \pm 2.66\%$ &$0$\\
			&CMDA& $86.38\pm1.62\%$ & $10,200$ \\
			&SimpleNet & $93.24 \pm 2.34\%$  & $2,040$\\
			&ReContrast (No pre-training)& $86.93 \pm 2.60\%$ & $1,600$ \\
			&ReContrast& $96.32 \pm 0.65\%$  & $1,600$\\
			\cline{2-4}
			&SAD+KKA (add)& $94.80 \pm 0.71\%$ &  $318$\\
			&SAD+KKA (rep)& $94.31 \pm 0.72\%$ &  $200$\\
			\cline{2-4}
			&SimpleNet+KKA (add)& $95.45 \pm 1.35\%$&$2,040 (318)$ \\
			&SimpleNet+KKA (rep)& $94.77 \pm 1.97\%$&$2,040 (200)$\\
			\cline{2-4}
			&ReContrast+KKA (add)   & $97.88 \pm 0.12\%$&$1,600+318$ \\
			&ReContrast+KKA (rep)  & $97.18 \pm 0.43\%$&$1,600+200$\\
			\hline
			\multirow{11}{*}{UCM-Caption} 
			& SSD & $71.89 \pm 2.68\%$ &$0$\\
			&CMDA& $72.06\pm1.22\%$ & $47,400$\\
			&SimpleNet & $83.80 \pm 1.60$\%&$6,320$ \\
			&ReContrast (No pre-training)& $72.62 \pm 0.94\%$ & $1,600$ \\
			&ReContrast& $79.46 \pm 0.69$\%&$1,600$ \\
			\cline{2-4}
			&SAD+KKA (add)& $74.98 \pm 0.49\%$ &  $318$\\
			&SAD+KKA (rep)& $74.60 \pm 0.70\%$ &  $200$\\
			\cline{2-4}
			&SimpleNet+KKA (add)& $94.01 \pm 0.91\%$&$6,320 (318)$ \\
			&SimpleNet+KKA (rep)& $94.77 \pm 1.97\%$&$6,320 (200)$\\
			\cline{2-4}
			&ReContrast+KKA (add)& $83.20 \pm 1.14\%$&$1,600+318$ \\
			&ReContrast+KKA (rep)& $82.67 \pm 2.04\%$&$1,600+200$\\
			\hline
		\end{tabularx}
		\caption{
			The results of KKA and baseline methods are presented across three datasets. The $\pm$ shows $95\%$ confidence interval over tasks. For KKA (add), we progressively inject new generated hard anomalies into the dataset with each iteration. For KKA (rep), the number of generated anomalies remains fixed, and we randomly replac easy anomalies in the dataset with hard anomalies.}
		\label{table: KK1}
	\end{table*}

	\section{Experiments}
	\label{sec:experiments}
	In this section, we present our experimental setup, evaluate the proposed method on multiple datasets, conduct ablation studies, perform hyperparameter analyses and convergence analysis. The code and data for the proposed method are provided for research purposes.
	
	\subsection{Datasets and Settings}
	
	We conduct experiments on three distinct datasets, each containing class labels that facilitate the classification of normal samples and anomalies.
	
	\noindent\textbf{\textit{CIFAR-100}} \cite{krizhevsky2009learning}. The CIFAR-100 image classification dataset comprises 20 superclasses with a total of 100 subclasses, each containing 600 images (500 for training and 100 for testing). Each image is assigned both a fine label and a coarse label. In this paper, we designate one superclass (comprising 5 subclasses) as normal samples, while the remaining 19 superclasses (95 subclasses) are considered anomalies.
	
	\noindent\textbf{\textit{Oxford-102}} \cite{reed2016generative}. 
	This dataset includes 8,189 image-text pairs of flowers across 102 distinct classes. In our experiments, one flower class is designated as the normal sample, while the remaining 101 classes are considered anomalies.
	
	\noindent\textbf{\textit{UCM-Caption}} \cite{qu2016deep}. This dataset comprises 21 classes of land-use images, with five distinct sentences used to describe each image. We randomly select two classes as normal samples, while the remaining 19 are designated as anomalies.

	\noindent\textbf{Implementation.} 
	In this paper, we focus on extracting anomaly-related knowledge and generating anomaly images for unsupervised anomaly detection. 
	
	The unsupervised baselines can be grouped into three categories: (1) methods based on raw data, such as SSD \cite{sehwagssd}; (2) methods that use cross-modal information to generate additional samples, such as CMDA; and (3) methods that produce anomalies, such as SimpleNet \cite{liu2023simplenet}. The proposed KKA belongs to the third category. 
	
	For KKA, the feature extractor’s learning rate is set to $0.0001$, with a scheduler adjustment at 50 epochs or 40 epochs. The batch size is 32, and the optimizer is Adam. The number of training epochs on the CIFAR-100, Oxford-102, and UCM-Caption datasets is 150, 200, and 300, respectively. Additionally, the generated anomaly dataset is updated 3, 2, and 2 times for CIFAR-100, Oxford-102, and UCM-Caption, respectively.
	
	
	\begin{table}[t]
		\centering
	\begin{tabularx}{\linewidth}{|c |X<{\centering} | X<{\centering}|}
			\hline
			Dataset & Methods & AUC \\
			\hline
			\multirow{5}{*}{CIFAR-100} & No anomalies & $52.48\pm0.48\%$ \\
			& KKA (0) & $70.18\pm0.85\%$ \\
			&KKA (1) & $71.46\pm1.39\%$ \\
			&KKA (2) & $71.97\pm1.29\%$ \\
			&KKA (3) & $72.26\pm1.19\%$ \\
			\hline
			\multirow{5}{*}{Oxford-102} 
			& No anomalies & $68.21 \pm 2.23\%$ \\
			&KKA (0) & $92.88 \pm 1.18\%$ \\
			&KKA (1) & $92.90 \pm 1.56\%$ \\
			&KKA (2) & $94.31 \pm 0.72\%$ \\
			&KKA (3) & $92.97 \pm 1.57\%$ \\
			\hline
			\multirow{5}{*}{UCM-Caption} 
			& No anomalies & $65.06 \pm 3.29\%$ \\
			&KKA (0) & $72.33 \pm 2.06\%$ \\
			&KKA (1) & $73.39 \pm 1.75\%$ \\
			&KKA (2) & $74.60 \pm 0.70\%$ \\
			&KKA (3) & $73.17 \pm 0.61\%$ \\
			\hline
		\end{tabularx}
		\caption{The results of SAD+KKA (rep) iteratively updating anomalies. The $\pm$ shows $95\%$ confidence interval over tasks. The number in parentheses represents the iteration count, with $0$ indicating the initially generated anomalies.}
		\label{table: KK2}
	\end{table}
	\subsection{Results}
	Table \ref{table: KK1} presents KKA's performance in both addition and replacement modes, along with the results of unsupervised methods with and without generated samples.
	KKA demonstrates strong generality, as it can be incorporated into supervised anomaly detection methods (such as SAD) as well as generation-based anomaly detection methods (such as SimpleNet). In general, introducing KKA leads to a significant improvement in detection performance.
	For instance, on the CIFAR-100 dataset, KKA improves SimpleNet’s AUC from $74.62\%$ to $84.04\%$ while generating only approximately $5\%$ of the samples produced by SimpleNet. Moreover, the superior performance of KKA based on SimpleNet demonstrates that extracting key knowledge from LLMs, rather than generating anomaly samples through random noise, is better suited for vision anomaly detection.
	The results of Recontrast \cite{guo2024recontrast} and ReContrast (No pre-training) clearly demonstrate the effectiveness of pre-training. In scenarios without anomalies, pre-trained models can integrate prior knowledge that can roughly help the detector understand what anomalies might be. Based on pre-trained model, Recontrast+KKA further introduces key knowledge specific to particular anomaly scenarios, resulting in additional performance improvements.
	Overall, the results in Table \ref{table: KK1} indicate that the samples generated by KKA outperform those generated by SimpleNet using random noise, ReContrast with multi-view samples, and CMDA with cross-modal information. This also demonstrates that knowledge extracted from LLMs is more practically relevant for specific anomaly detection scenarios.
	
		\begin{figure*}[t]
		\centering
		\includegraphics[scale=0.52]{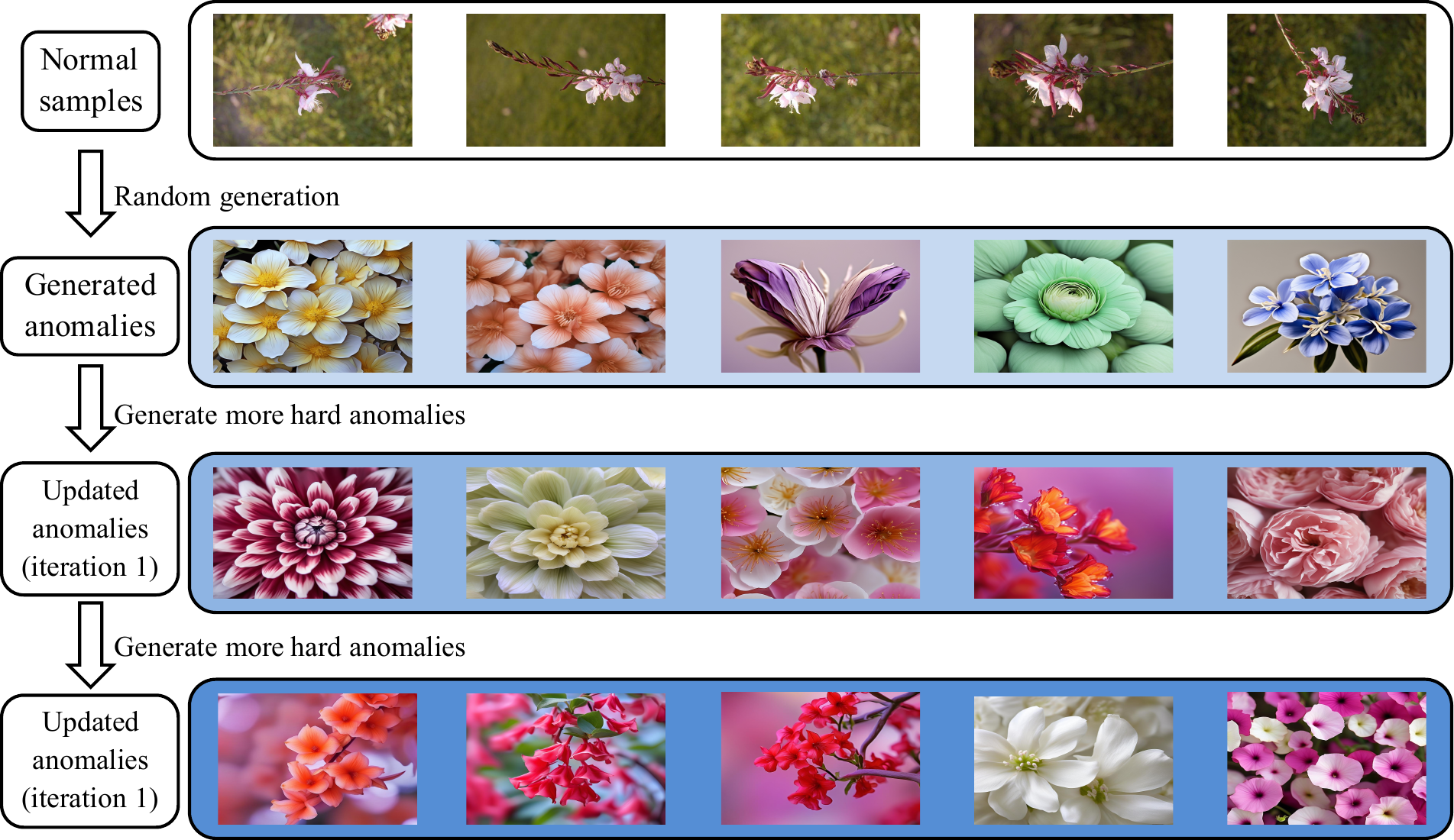}
		\caption{The anomalies generated by KKA on Oxford-102 dataset and the corresponding iterative process.}
		\label{Fig: Generated image}
	\end{figure*}

	\begin{figure}[t]
		\centering
		\includegraphics[scale=0.6]{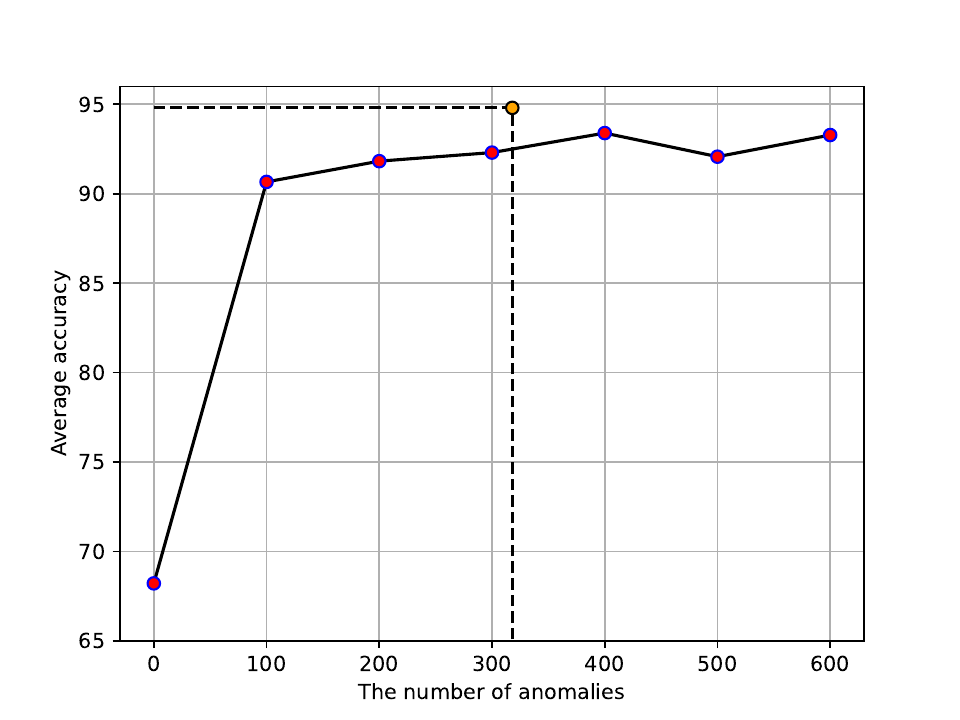}
		\caption{Different quantities of initially generated anomalies and their corresponding average accuracy on Oxford-102.}
		\label{Fig: HA}
	\end{figure}

	\subsection{Importance of Key Knowledge}
	
	We conduct ablation experiments to demonstrate the importance of key knowledge (hard anomalies). 
	To alleviate concerns that performance improvements may be due to an increase in data volume, we present the results of KKA (replacement). KKA (replacement) maintains a fixed number of anomalies while gradually increasing the proportion of hard anomalies with each iteration.
	The related results are presented in Table \ref{table: KK2}, where (0) denotes the initially generated anomalies, and (1), (2), and (3) represent the first, second, and third iterations of the updated anomaly dataset, with higher numbers indicating a greater proportion of hard anomalies.

	Related results indicate that increasing the proportion of hard anomalies (1, 2, 3), can significantly enhance detector's performance. For CIFAR-100, the AUC consistently improves with each iteration. In contrast, while detector's performance also improves on the Oxford-102 and UCM-Caption datasets, it is less stable due to differences in data volume: CIFAR-100 includes 2500 normal samples, whereas Oxford-102 and UCM-Caption contain only 51 and 158 normal samples, respectively. An excessive number of hard anomalies may cause the detector to overlook the fewer normal samples. Meanwhile, on Oxford-102 and UCM-Caption, the AUC for KKA (0) is $92.88\%$ and $72.33\%$, while the detector achieves $94.31\%$ and $74.60\%$ on the second iteration. The improvement in detector's performance achieved by increasing the proportion of hard anomalies clearly demonstrates the importance of key knowledge (hard anomalies) in anomaly detection.

	Figure \ref{Fig: HA} further illustrates the relationship between the number of generated anomalies and the number of hard anomalies. In Figure \ref{Fig: HA}, orange points represent KKA applied to increase hard anomalies, while pink points represent the different number of initial anomalies (including more easy anomalies). The results show that increasing the number of initial anomalies has little effect on the detector's performance. Moreover, once the number of generated anomalies reaches 400, the detector's performance plateaus. Datasets with a higher proportion of hard anomalies can easily outperform those with a larger number of easy anomalies. This indicates that hard anomalies contain more meaningful information for anomaly detection.

	\subsection{Visualization of KKA}
	We demonstrate the iterative process of generating anomalies by KKA on the Oxford-102 dataset, and the results are presented in Figure \ref{Fig: Generated image}. 
	
	In the initial generation, the anomalies exhibit high randomness, featuring flowers of various colors and shapes extracted from the prior knowledge of LLMs, ensuring the diversity of anomalies. 
	In the first iteration, guided by the selected hard anomalies, KKA extracts key knowledge from the prior information of LLMs, making the anomalies more difficult to distinguish from normal samples. As a result, the generated images in iteration 1 contain elements that resemble normal samples, such as smaller flowers and a pink coloration. In the second iteration, the influence of key knowledge becomes more prominent in the generated anomalies. Although the anomalies remain distinguishable from normal samples, their characteristics increasingly resemble those of normal samples, with a higher proportion than in the previous iteration.

\begin{figure}[t]
		\centering
		\subfloat[Latent space learned by SSD]{\includegraphics[width=3.5in]{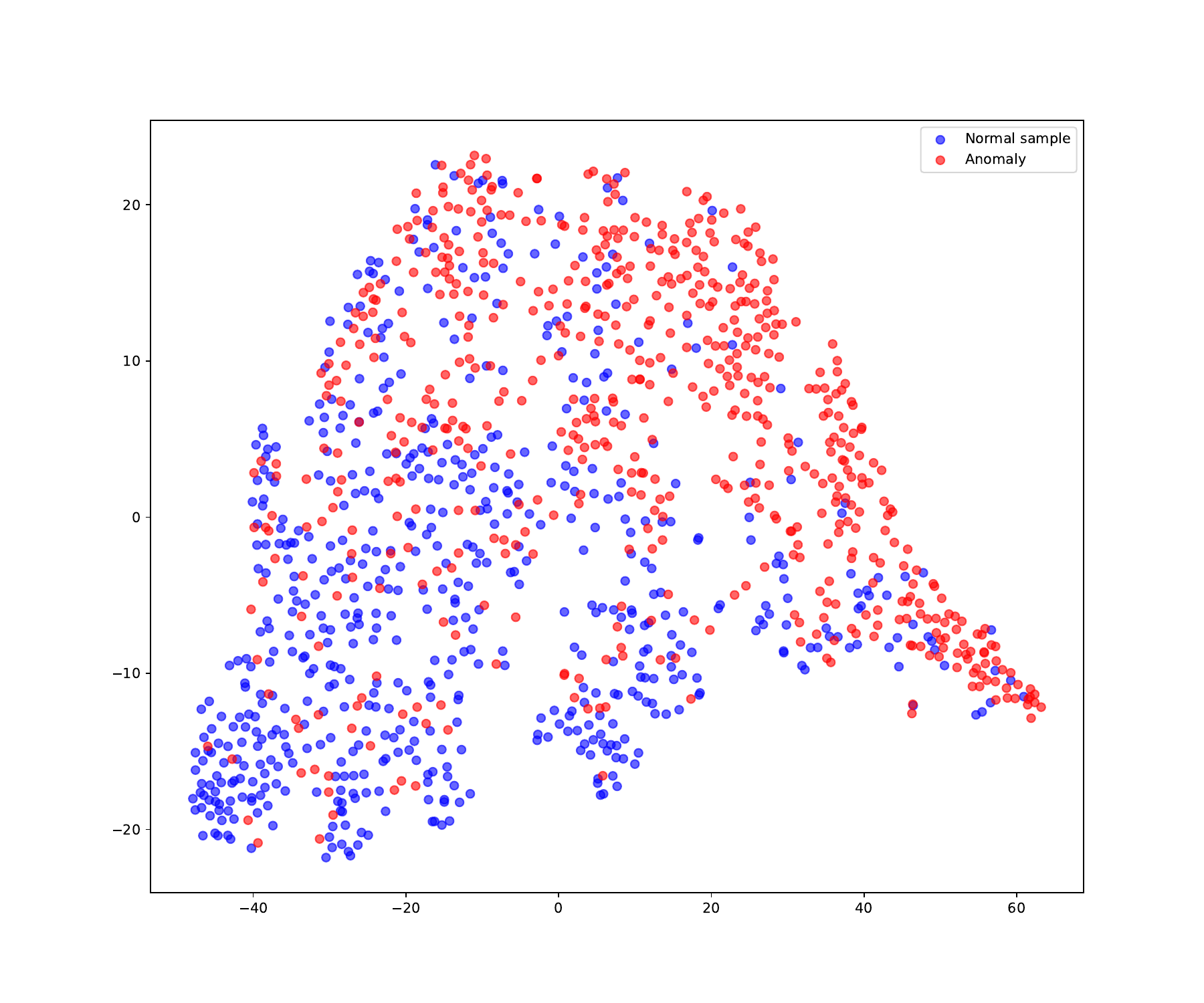}%
			\label{fig:sub1}}
		\hfil
		\subfloat[Latent space learned by KKA]{\includegraphics[width=3.5in]{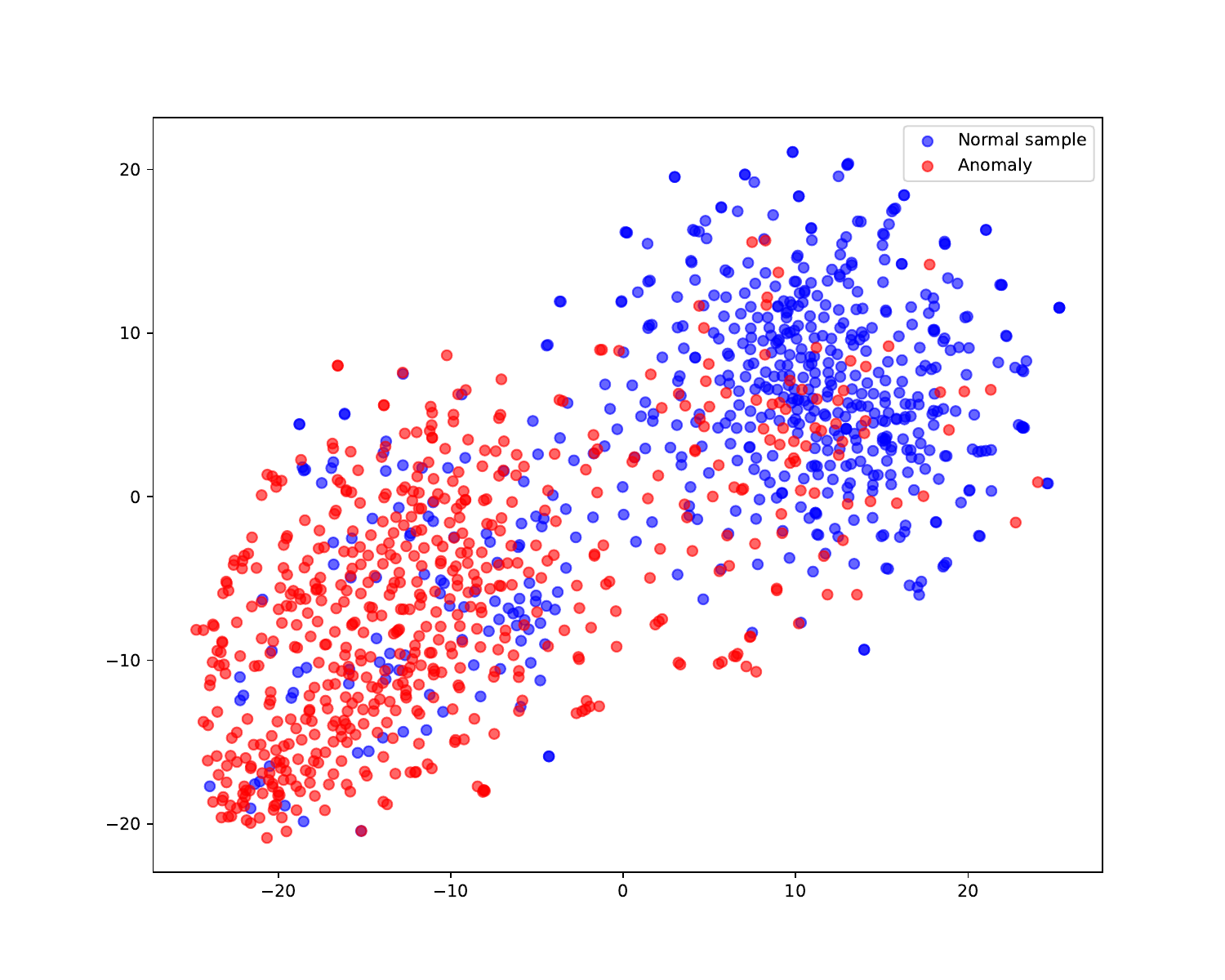}%
			\label{fig:sub2}}
		\caption{Visualization of different methods. The dimensionality-reduction algorithm for visualizing high-dimensional samples is t-SNE \cite{van2008visualizing}.}
		\label{Fig: tsne}
	\end{figure}

	We visualize the latent spaces learned by the unsupervised methods, SSD, and the proposed KKA in Figure \ref{Fig: tsne} to demonstrate that the proposed method learns more distinct boundaries between normal samples and anomalies.

	As illustrated in Figure \ref{fig:sub1}, the latent space learned by SSD is more chaotic. In contrast, in Figure \ref{fig:sub2}, the latent space learned by KKA can clearly separate most normal samples and anomalies, as there are clear distribution differences between blue dots and red dots. More specifically, while SSD is able to project some of the red points to relatively concentrated areas (i.e., easy anomalies), there are still a large number of red points interspersed with the blue points (i.e., hard anomalies). 
	
		\begin{figure}[t]
		\centering
		\includegraphics[scale=0.57]{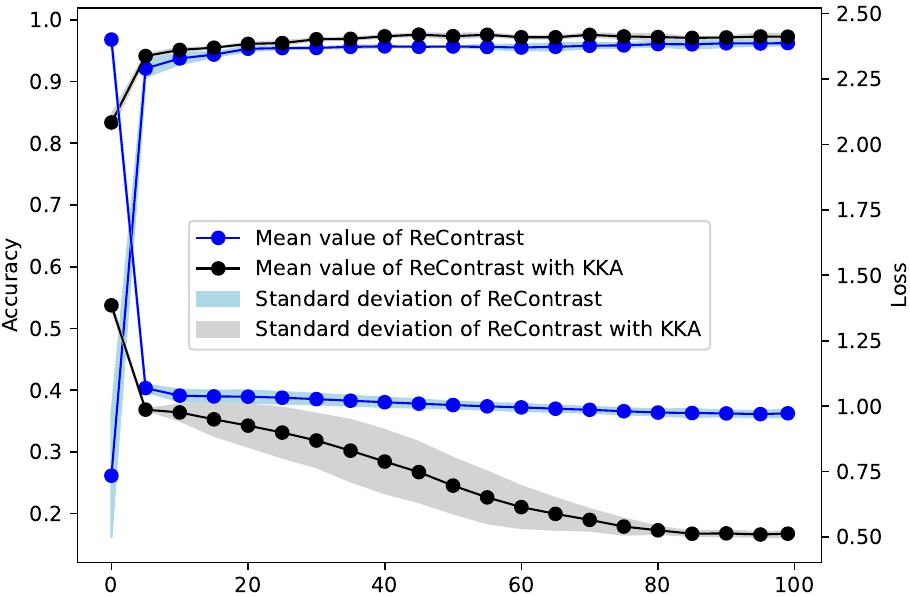}
		\caption{Convergence analysis of ReContrast and ReContrast with KKA on Oxford-102.}
		\label{Fig: convergence}
	\end{figure}

	\subsection{Different Setting for Anomaly Detection}
	\begin{table}[t]
		\centering
	\begin{tabularx}{\linewidth}{|c| X<{\centering}|  X<{\centering}|}
			\hline
			Dataset & Methods & AUC (\%) \\
			\hline
			\multirow{5}{*}{CIFAR-100} & No anomalies & $72.95 \pm 2.40\%$ \\
			&KKA (0) & $78.51 \pm 5.02\%$ \\
			&KKA (1) & $80.46 \pm 4.41\%$ \\
			&KKA (2) & $81.79 \pm 2.11\%$ \\
			&KKA (3) & $82.18 \pm 3.19\%$ \\
			\hline
		\end{tabularx}
		\caption{More experiments on CIFAR-100, where normal samples from one subclass, anomalies from 99 subclasses.}
		\label{table: KK3}
	\end{table}
	In real-world scenarios, anomalies are inherently difficult to predict. They may not follow patterns like those in the Flower-102 or UCM datasets, where normal samples and anomalies generally belong to the same class. 
	Instead, anomalies are more likely to originate from two aspects: similar distributions (i.e., the same superclass but different subclasses) and distinct distributions (i.e., different superclasses). 
	To further validate the effectiveness of KKA, we restructured the CIFAR-100 dataset by selecting one class as the normal samples and treating the remaining 99 classes as anomalies. The results are presented in Table \ref{table: KK3}.
	It is evident that when only one subclass is used as normal samples, the task is relatively easier compared to using a superclass (comprising five subclasses) as normal samples. This is because the distribution of normal samples is more homogeneous, allowing the detector to more focus on their distribution. Furthermore, the results of KKA in Tables \ref{table: KK1} and \ref{table: KK3} exhibit a consistent pattern: as the number of iterations increases, the detector's performance improves. This further underscores the important role of hard anomalies in anomaly detection.

	\subsection{Convergence Analysis}
	In Figure \ref{Fig: convergence}, we demonstrate that the anomalies generated by KKA consistently enhance the performance of the best-performing baseline, ReContrast. Notably, all settings for ReContrast and ReContrast with KKA are identical. As illustrated in Figure \ref{Fig: convergence}, with the more realistic anomaly samples generated by KKA, ReContrast can converge faster and more effectively. This is because the anomalies are no longer random, unrealistic content but rather real-world images in the same "style" as the normal samples. 
	Besides, during the first epoch, the key knowledge extracted by KKA increased ReContrast's accuracy from approximately 0.25 to 0.83, while the loss decreased by nearly 1. Such results highlight the crucial role of hard anomalies generated from key knowledge in improving detector's performance.

	\section{Conclusion}
	Unsupervised anomaly detection, particularly in the visual domain, faces a significant challenge in distinguishing normal samples from diverse anomalies. Existing generation-based anomaly detection methods alleviate such issue by generating anomalies that deviate from the normal distribution. However, generated samples lacking real-world knowledge guidance are often unrealistic and overly broad, lacking specificity with respect to normal samples. 
	This paper proposes extracting key knowledge about anomalies from the extensive prior knowledge of LLMs and generating anomaly images. To overcome the limitations of random anomaly generation, we categorize the generated anomalies into easy anomalies and hard anomalies according to their level of confusion by a trained feature extractor. By iteratively increasing the proportion of hard anomalies, the proposed method can further improve the detector's effectiveness.

\section*{Acknowledgments}
This work was supported by the National Key Research and Development Program of China (No.2021YFB3301504), the Excellent Young Scientist Fund of Henan Province (No. 32300421095), China Postdoctoral Special Funding Program (No. 2022T150590).
	

	\bibliographystyle{IEEEtran}
	\bibliography{IEEEabrv,sample-base}

\end{document}